\documentclass[10pt, conference]{IEEEtran}

\usepackage{graphicx}
\usepackage{float}
\usepackage[font=small,labelfont=bf,skip=5pt,labelsep=space]{caption}
\usepackage{amsmath}
\usepackage{booktabs}
\usepackage{algorithm}
\usepackage{algpseudocode}
\usepackage[colorlinks=true, linkcolor=blue, urlcolor=blue]{hyperref}
\usepackage{balance}
\usepackage{etoolbox}

\begin{document}

\title{\LARGE \bf Remaining Useful Life Estimation for Turbofan Engines: A Comparative Study of Classical, CNN, and LSTM Approaches}

\author{
\IEEEauthorblockN{Galchar Samarthkumar, Goel Astitva, Kanu Sumit}
\IEEEauthorblockA{Khoury College of Computer Science, Northeastern University\\
galchar.sa@northeastern.edu, goel.ast@northeastern.edu, kanu.s@northeastern.edu}
}

\maketitle

\begin{abstract}
 Remaining Useful Life (RUL) estimation is a critical component of Prognostics and Health Management (PHM), enabling proactive maintenance scheduling and reducing unplanned failures in industrial equipment. This paper presents a comparative study of machine learning approaches for RUL estimation on the NASA C-MAPSS turbofan engine dataset: classical baselines (Ridge Regression, Polynomial Ridge, and XGBoost), a 1D Convolutional Neural Network (CNN), and a Long Short-Term Memory (LSTM) network. All models are evaluated on the FD001 and FD003 subsets under an identical preprocessing pipeline to ensure a fair comparison. Among raw-sequence models, the LSTM achieves RMSE of 14.93 and 14.20 on FD001 and FD003 respectively, outperforming the deep LSTM reported by Zheng et al.~\cite{paper} (RMSE 16.14 and 16.18) despite using a simpler single-layer architecture. The 1D CNN achieves RMSE of 16.97 on FD001 and 15.68 on FD003, demonstrating competitive performance on FD003 while producing more conservative RUL predictions on FD001. Ridge Regression is evaluated on raw and engineered features, while other classical models use only engineered inputs. XGBoost achieves an RMSE of 13.36 on FD003, highlighting the competitiveness of nonlinear modeling.
\end{abstract}

\section{Introduction}

Predictive maintenance has emerged as a key strategy in modern industrial operations, moving away from costly scheduled maintenance toward condition-based interventions. At its core lies the problem of RUL estimation: predicting how many operational cycles remain before a component fails. Accurate RUL estimation allows engineers to schedule maintenance in advance, reducing unnecessary downtime and preventing catastrophic failures. In high-stakes domains such as aviation, power generation, and mining, inaccurate RUL predictions can have severe safety and economic consequences.

Data-driven approaches to RUL estimation have been prevalent due to the widespread availability of sensor data from industrial equipment. These approaches learn degradation patterns directly from historical run-to-failure data, without requiring explicit physical failure models. The NASA C-MAPSS dataset~\cite{cmapss} is one of the most widely used benchmarks for evaluating such methods, providing multivariate time-series sensor data from simulated turbofan engines under various operating and fault conditions.

This work compares multiple machine learning paradigms for RUL estimation under a strictly controlled experimental setup. For classical models, we evaluate \textbf{Ridge Regression} on both raw flattened sliding-window inputs and engineered statistical features, while \textbf{Polynomial Ridge Regression and XGBoost} are applied to the engineered features. On the deep learning side, we evaluate a \textbf{1D CNN} that treats windowed sensor data as a spatial signal, and an \textbf{LSTM network} that explicitly models temporal dependencies across the full sensor sequence. The deep learning models are evaluated on raw sensor sequences for a fair architectural comparison, while classical models are evaluated on both raw flattened windows and engineered statistical features, with the latter serving as a strong baseline for achievable performance on this dataset.

\section{Background}

RUL estimation methods fall into two broad categories: model-based and data-driven. Model-based approaches construct physical degradation models and are useful when failure data is scarce, but require deep domain expertise and are difficult to generalize. Data-driven methods learn from historical sensor observations and are more scalable in data-rich environments.

Early data-driven approaches used sliding window features fed into shallow methods such as Support Vector Regression (SVR) and Multi-Layer Perceptrons (MLP)~\cite{paper}. These methods represent the sensor sequence as a fixed-size feature vector, discarding temporal ordering.

Hidden Markov Models (HMMs) were among the first sequence learning approaches applied to prognostics~\cite{paper}, but their discrete state space and Markov assumption limit capacity for complex, long-range dependencies. Recurrent Neural Networks (RNNs) can model sequences but suffer from the vanishing gradient problem~\cite{bengio1994}, which prevents effective learning over long horizons.

Long Short-Term Memory networks, introduced by Hochreiter and Schmidhuber~\cite{lstm}, address this through a gating mechanism comprising input, forget, and output gates that selectively retain and discard information over time. Zheng et al.~\cite{paper} demonstrated that a deep LSTM outperforms CNN and shallow methods on C-MAPSS, particularly on the NASA scoring function which penalizes overestimating RUL more than underestimating it.

1D CNNs have also been applied to RUL estimation by treating windowed sequences as spatial signals~\cite{babu2016}. They are computationally efficient and effective at capturing local temporal patterns, but lack an explicit memory mechanism for long-range sequence dependencies.

\section{Approach}

\subsection{Dataset}

We use the NASA C-MAPSS dataset~\cite{cmapss}, specifically the FD001 and FD003 subsets. Each time step contains readings from 21 sensors and 3 operational settings. Table~\ref{tab:dataset} summarizes the characteristics of both subsets.

\begin{table}[t]
\caption{C-MAPSS Dataset Characteristics}
\label{tab:dataset}
\centering
\begin{tabular}{lcccc}
\toprule
\textbf{Subset} & \textbf{Train} & \textbf{Test} & \textbf{Op. Cond.} & \textbf{Fault Cond.} \\
\midrule
FD001 & 100 & 100 & 1 & 1 \\
FD003 & 100 & 100 & 1 & 2 \\
\bottomrule
\end{tabular}
\end{table}

\subsection{Preprocessing}

All models share an identical preprocessing pipeline.

\textbf{RUL Target Function:} Through EDA, we observed that engines exhibit a stable healthy phase before meaningful degradation begins. Based on this, we apply a piece-wise linear RUL target capped at 130 cycles. Early healthy cycles are assigned a constant RUL of 130, preventing the model from learning spurious degradation signals during stable operation.

\textbf{Sensor Selection:} Sensors with near-zero variance carry no degradation signal. We remove \texttt{s1, s5, s10, s16, s18, s19} for both subsets, and additionally \texttt{s6} for FD001 only, yielding 14 sensors for FD001 and 15 for FD003. This decision is grounded in EDA: the dropped sensors exhibit constant or near-constant readings across all engine lifecycles, while \texttt{s6} is retained for FD003 where it shows a correlation of $-0.215$ with RUL and nonlinear spread near failure due to the multiple fault conditions.

\textbf{Normalization:} All sensor readings are z-score normalized. The scaler is fit on training data only to prevent leakage into validation and test sets.

\textbf{Train/Validation Split:} Engines are split 80/20 by engine ID to ensure the model is evaluated on entirely unseen degradation trajectories.

\textbf{Sliding Window:} A sliding window of 30 cycles with stride 1 generates training sequences, each labeled with the RUL at the final time step. For test data, the last 30 cycles of each engine are used, with zero-padding for engines shorter than 30 cycles.

\subsection{Linear and Tree-Based Baselines}
The Ridge Regression baseline is evaluated using two input representations. In the raw setting, each 30-cycle window is flattened into a vector and fed directly to the model. In the feature-engineering setting, five features are extracted per sensor from each window: mean, standard deviation, last observed value, delta (last minus first), and linear slope. This yields 70 features per window for FD001 and 75 for FD003. These are fed into a Ridge model with $L_2$ regularization. While the engineered representation is unable to accommodate temporal ordering, the slope and delta features compensate by encoding degradation direction.
\begin{algorithm}[h!]
\caption{Ridge Regression for RUL Estimation}
\begin{algorithmic}[1]
    \Require C-MAPSS dataset $D$, mode $\in \{\text{raw}, \text{engineered}\}$
    \Ensure Predicted RUL $\hat{y}$ per test engine
    \For{each training engine $e$ in $D$}
        \State Compute per-cycle RUL labels, clip at $\text{MAX\_RUL} = 130$
        \For{each window $W$ of size 30}
            \If{mode = engineered}
                \State $F \gets [\text{mean}, \text{std}, \text{last}, \text{delta}, \text{slope}]$ per sensor
            \Else
                \State $F \gets \text{flatten}(W)$
            \EndIf
            \State Append $(F,\ \text{RUL at last step})$ to training set
        \EndFor
    \EndFor
    \State Normalize $F$ using z-score normalization (fit on train only)
    \State Train Ridge: $\hat{y} = Fw + b$ with $L_2$ penalty ($\alpha = 1.0$)
    \State For test: extract last window per engine, apply same transformation, predict $\hat{y}$
\end{algorithmic}
\end{algorithm}
Polynomial Ridge expands the engineered feature vector with degree-2 interaction and squared terms, capturing nonlinear relationships between sensor statistics.
\begin{algorithm}[h!]
\caption{Polynomial Ridge Regression for RUL Estimation}
\begin{algorithmic}[1]
    \Require C-MAPSS dataset $D$
    \Ensure Predicted RUL $\hat{y}$ per test engine
    \State Extract engineered features $F$ as in Algorithm 1 (mode = engineered)
    \State Normalize $F$ using z-score normalization (fit on train only)
    \State $F_{\text{poly}} \gets \text{PolynomialFeatures}(F,\ \text{degree}=2)$
    \State Train Ridge: $\hat{y} = F_{\text{poly}}w + b$ with $L_2$ penalty ($\alpha = 1.0$)
    \State For test: extract last window per engine, apply same transformations, predict $\hat{y}$
\end{algorithmic}
\end{algorithm}
XGBoost uses gradient-boosted decision trees on the same engineered features, capturing complex nonlinear interactions without requiring polynomial expansion.
\begin{algorithm}[h!]
\caption{XGBoost for RUL Estimation}
\begin{algorithmic}[1]
    \Require C-MAPSS dataset $D$
    \Ensure Predicted RUL $\hat{y}$ per test engine
    \State Extract engineered features $F$ as in Algorithm 1 (mode = engineered)
    \State Train XGBoost regressor:
    \State \hspace{1em} $n\_estimators = 500$, $max\_depth = 6$, $learning\_rate = 0.05$
    \State \hspace{1em} $subsample = 0.8$, $colsample\_bytree = 0.8$
    \State \hspace{1em} Early stopping on validation set (patience $= 20$)
    \State For test: extract last window per engine and predict $\hat{y}$
\end{algorithmic}
\end{algorithm}

\subsection{1D Convolutional Neural Network}

The CNN treats each windowed sensor sequence as a 1D signal with sensors as channels. Three convolutional layers with kernel size 3 and same padding extract local temporal features across the 30-cycle window, progressively increasing filter depth from $n$ to 32 to 64. Dropout ($p=0.3$) is applied after the first two convolutional layers for regularization. The resulting feature maps are flattened and passed through a two-layer fully connected block mapping to a scalar RUL prediction with a linear output activation, appropriate for unconstrained regression.

The architecture was validated through an ablation study comparing a deeper four-layer variant with varying kernel sizes (3, 5, 7, 3) and BatchNorm. The deeper model underperformed on both datasets, confirming that with only 100 training engines, regularization is more beneficial than increased model capacity.
\begin{algorithm}[h!]
\caption{1D CNN for RUL Estimation}
\begin{algorithmic}[1]
    \Require C-MAPSS sensor dataset $D$
    \Ensure Predicted RUL $\hat{y}$ per test engine
    \For{each engine $e$ in $D$}
        \State Compute RUL labels, clip at $\text{MAX\_RUL} = 130$
        \For{each sliding window $W$ of size 30}
            \State $X \gets W^{\top}$ \Comment{transpose to (sensors, window)}
            \State Append $(X,\ \text{RUL at last step})$ to training set
        \EndFor
    \EndFor
    \State Normalize sensors using z-score (fit on train only)
    \For{each training batch $(X_b, y_b)$}
        \State $F \gets \text{Conv1D}(n \to 32) \to \text{ReLU} \to \text{Dropout}(0.3)$
        \State $F \gets \text{Conv1D}(32 \to 64) \to \text{ReLU} \to \text{Dropout}(0.3)$
        \State $F \gets \text{Conv1D}(64 \to 64) \to \text{ReLU}$
        \State $\hat{y}_b \gets \text{Flatten} \to \text{Linear}(64{\times}30 \to 128) \to \text{ReLU} \to \text{Dropout}(0.3) \to \text{Linear}(128 \to 1)$
        \State Update weights: $\mathcal{L} = \text{MSE}(\hat{y}_b, y_b)$
    \EndFor
    \State For test: extract last window per engine, predict $\hat{y}$
\end{algorithmic}
\end{algorithm}

\subsection{LSTM}
The LSTM processes each sensor sequence in temporal order. The final architecture was determined through an ablation study on layer depth and regularization, yielding a single LSTM layer followed by a three-layer fully connected block.
\begin{algorithm}[h!]
\caption{LSTM for RUL Estimation}
\begin{algorithmic}[1]
    \Require C-MAPSS sensor dataset $D$
    \Ensure Predicted RUL $\hat{y}$ per test engine
    \For{each engine $e$ in $D$}
        \State Compute RUL labels, clip at $\text{MAX\_RUL} = 130$
        \For{each sliding window $W$ of size 30}
            \State Append $(W,\ \text{RUL at last step})$ to training set
        \EndFor
    \EndFor
    \State Normalize sensors using z-score (fit on train only)
    \For{each training batch $(X_b, y_b)$}
        \State $h_1, \ldots, h_T \gets \text{LSTM}(n \to 32,\ X_b)$
        \State $h_T \gets \text{Dropout}(0.5,\ h_T)$
        \State $\hat{y}_b \gets \text{Linear}(32 \to 8) \to \text{ReLU} \to \text{Linear}(8 \to 8) \to \text{ReLU} \to \text{Linear}(8 \to 1)$
        \State Update weights: $\mathcal{L} = \text{MSE}(\hat{y}_b, y_b)$
    \EndFor
    \State For test: pad/trim to last 30 cycles per engine, predict $\hat{y}$
\end{algorithmic}
\end{algorithm}
Only the hidden state at the final time step $h_T$ is passed to the fully connected block, as it encodes the full sequence history.

\subsection{Training Protocol}

The CNN and LSTM are trained under identical conditions. Both models use RMSprop with a learning rate of 0.001 and weight decay of $10^{-5}$, optimizing MSE loss with a batch size of 64. A ReduceLROnPlateau scheduler (factor 0.5, patience 5) reduces the learning rate when validation loss plateaus, and training is stopped early if validation loss does not improve for 20 consecutive epochs (max 200 epochs). For classical models, Ridge and Polynomial Ridge are trained using $L_2$ regularization with $\alpha = 1.0$. XGBoost is trained using early stopping on the validation set, with a maximum of 500 trees and a patience of 20 rounds. All experiments use random seed 42 for reproducibility.

\subsection{Evaluation Metrics}

We evaluate all models on four metrics. \textbf{RMSE} and \textbf{MAE} measure prediction error magnitude. \textbf{R\textsuperscript{2}} measures variance explained by the model.

The \textbf{NASA Scoring Function}~\cite{paper} applies an asymmetric penalty, more heavily penalizing overestimation of RUL since late maintenance predictions risk equipment failure:
\begin{equation}
    S = \sum_{i=1}^{n} \begin{cases} e^{-h_i/13} - 1 & h_i < 0 \\ e^{h_i/10} - 1 & h_i \geq 0 \end{cases}, \quad h_i = \hat{y}_i - y_i
\end{equation}
Lower values indicate better performance for all metrics.

\section{Results}
\label{sec:results}

Table~\ref{tab:results} presents test set results for all models on FD001 and FD003. For completeness, Table~\ref{tab:mlp_extended} reports additional linear model variants using engineered features on FD003.

\begin{table}[h]
\caption{Extended Linear Model Results on FD003\\ (Engineered Features)}
\label{tab:mlp_extended}
\centering
\begin{tabular}{lcccc}
\toprule
\textbf{Model} & \textbf{RMSE} & \textbf{MAE} & \textbf{R\textsuperscript{2}} & \textbf{NASA} \\
\midrule
Ridge  & 17.494 & 14.398 & 0.809 & 531.1 \\
Polynomial Ridge         & 16.604 & 12.876 & 0.828 & 630.9 \\
XGBoost                  & \textbf{13.355} & \textbf{9.604} & \textbf{0.889} & \textbf{396.7} \\
\bottomrule
\end{tabular}
\end{table}

\begin{table*}[t]
\caption{Test Results on FD001 and FD003}
\label{tab:results}
\centering
\begin{tabular}{llcccccccc}
\toprule
& & \multicolumn{4}{c}{\textbf{FD001}} & \multicolumn{4}{c}{\textbf{FD003}} \\
\cmidrule(lr){3-6} \cmidrule(lr){7-10}
\textbf{Model} & \textbf{Type} & \textbf{RMSE} & \textbf{MAE} & \textbf{R\textsuperscript{2}} & \textbf{NASA} & \textbf{RMSE} & \textbf{MAE} & \textbf{R\textsuperscript{2}} & \textbf{NASA} \\
\midrule
Raw Ridge Regression & Linear    & 16.634 & 13.570 & 0.833 & 469.7 & 17.981 & 15.160 & 0.799 & 622.5 \\
1D CNN           & Deep CNN  & 16.971 & 12.868 & 0.826 & \textbf{365.0} & 15.678 & 11.299 & 0.847 & 647.6 \\
LSTM      & Deep LSTM & \textbf{14.931} & \textbf{11.732} & \textbf{0.865} & 410.97 & \textbf{14.204} & \textbf{10.612} & \textbf{0.874} & \textbf{409.1} \\
\midrule
Zheng et al.~\cite{paper} & Deep LSTM & 16.14 & {\textemdash} & {\textemdash} & 338 & 16.18 & {\textemdash} & {\textemdash} & 852 \\
\bottomrule
\end{tabular}
\end{table*}

\subsection{Training Curves}

\begin{figure*}[t]
    \centering
    \includegraphics[width=0.48\textwidth]{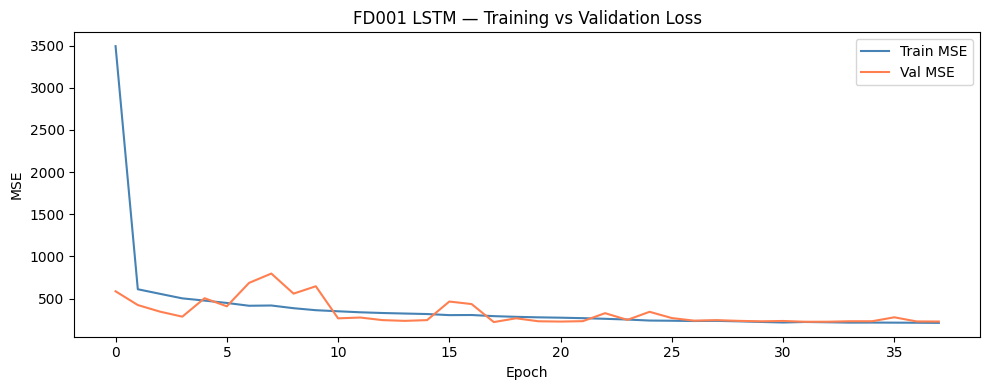}
    \hfill
    \includegraphics[width=0.48\textwidth]{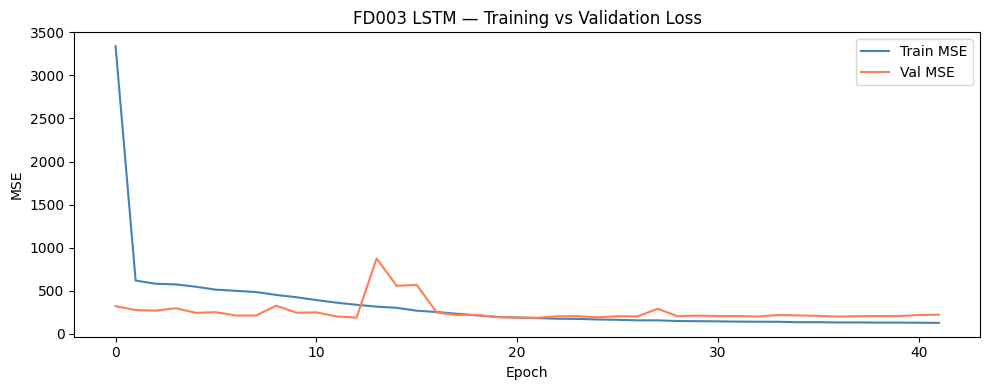}
    \caption{LSTM training and validation MSE loss over epochs for FD001 (left) and FD003 (right).}
    \label{fig:loss_curves_lstm}
\end{figure*}

Figure~\ref{fig:loss_curves_lstm} shows the training and validation loss curves for the LSTM on both datasets. The LR scheduler reduces the learning rate when validation loss plateaus, enabling continued improvement before early stopping triggers.

\begin{figure*}[t]
    \centering
    \includegraphics[width=0.48\textwidth]{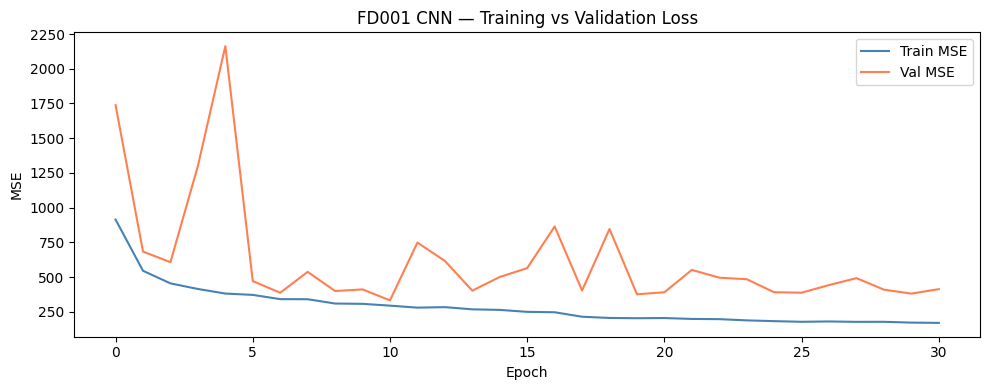}
    \hfill
    \includegraphics[width=0.48\textwidth]{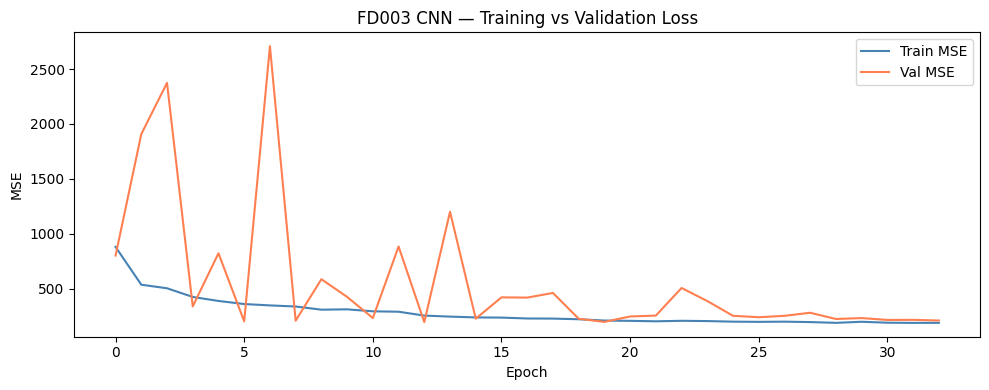}
    \caption{CNN training and validation MSE loss over epochs for FD001 (left) and FD003 (right). High initial validation spikes stabilize rapidly as the convolutional filters adapt to sensor degradation patterns.}
    \label{fig:loss_curves_cnn}
\end{figure*}

Figure~\ref{fig:loss_curves_cnn} shows the corresponding curves for the CNN. Both datasets exhibit high initial validation loss followed by rapid convergence, with the characteristic spikes typical of RMSprop optimization on small datasets. The CNN converges within approximately 30 epochs on both subsets, with early stopping restoring the best validation checkpoint.

\subsection{Predicted vs. True RUL}

\begin{figure*}[t]
    \centering
    \includegraphics[width=0.48\textwidth]{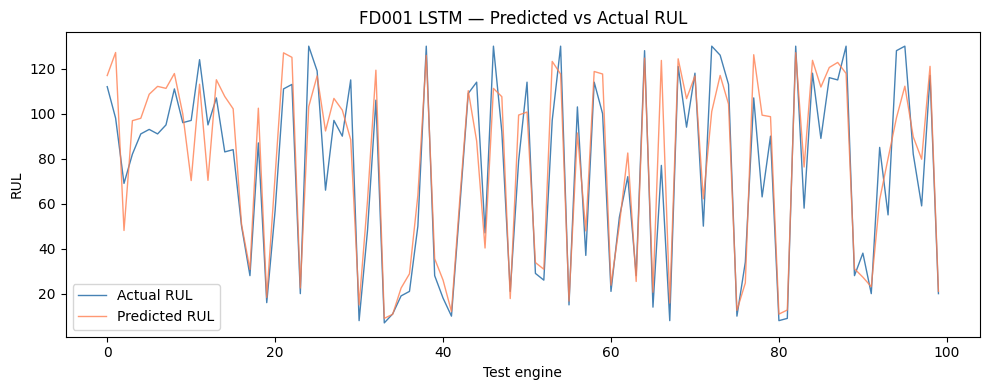}
    \hfill
    \includegraphics[width=0.48\textwidth]{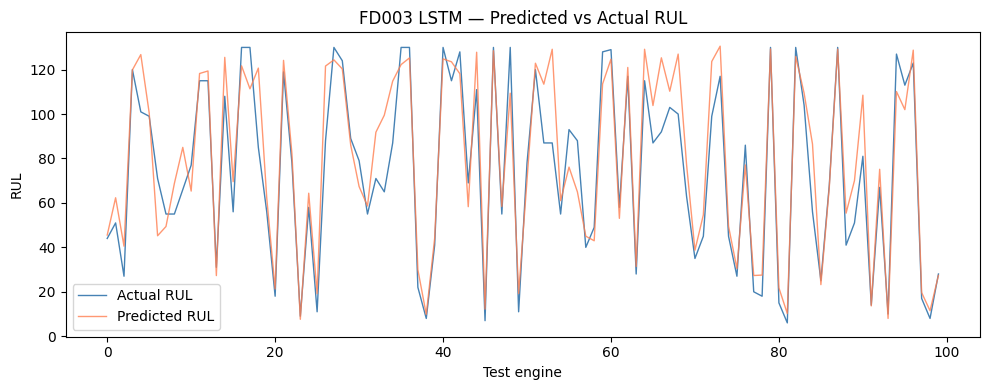}
    \caption{Predicted versus true RUL for the LSTM on the FD001 (left) and FD003 (right) test sets.}
    \label{fig:predictions_lstm}
\end{figure*}

\begin{figure*}[t]
    \centering
    \includegraphics[width=0.48\textwidth]{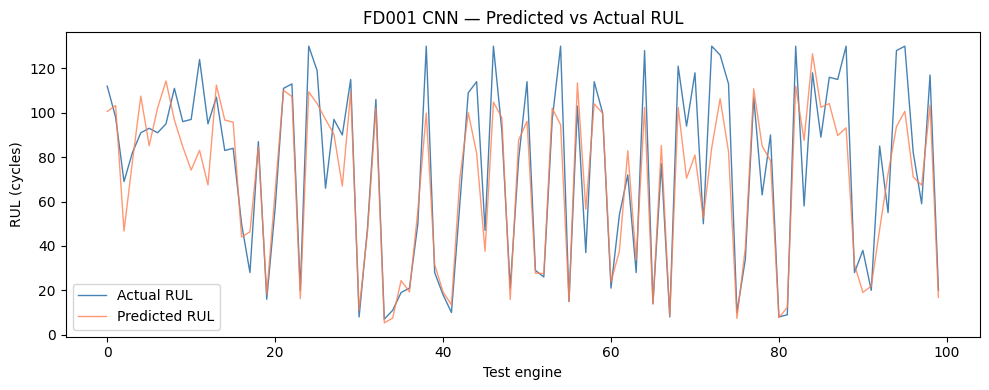}
    \hfill
    \includegraphics[width=0.48\textwidth]{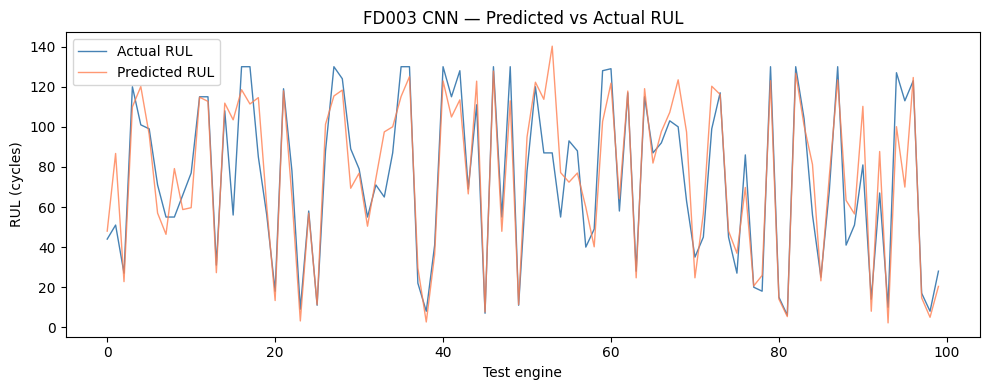}
    \caption{Predicted versus true RUL for the CNN on the FD001 (left) and FD003 (right) test sets.}
    \label{fig:predictions_cnn}
\end{figure*}

Figure~\ref{fig:predictions_lstm} shows predicted versus true RUL on the test sets for the LSTM. Figure~\ref{fig:predictions_cnn} shows the corresponding plots for the CNN. On FD001, the CNN predictions track the actual RUL closely across most engines, with the primary errors concentrated at sharp peaks where the model slightly underestimates high RUL values. On FD003, the tracking remains strong but with greater spread, consistent with the heterogeneous degradation patterns introduced by two fault modes.

\begin{figure*}[t]
    \centering
    \includegraphics[width=0.48\textwidth]{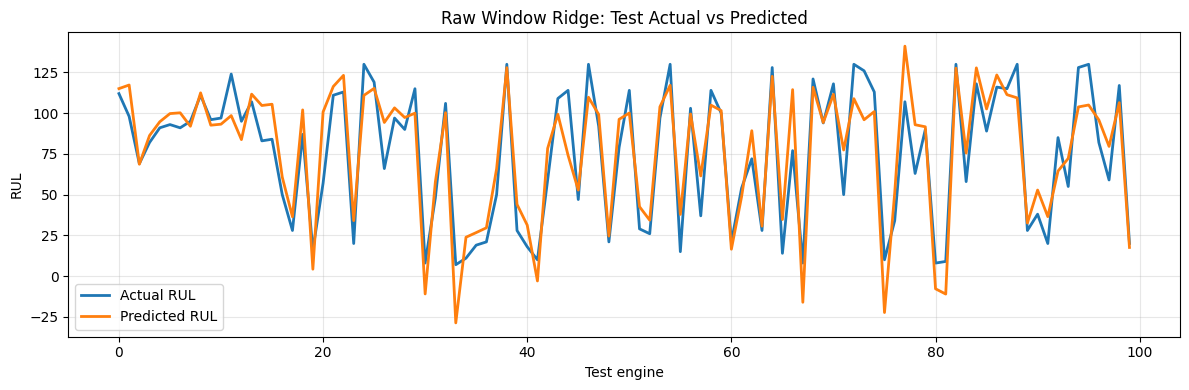}
    \hfill
    \includegraphics[width=0.48\textwidth]{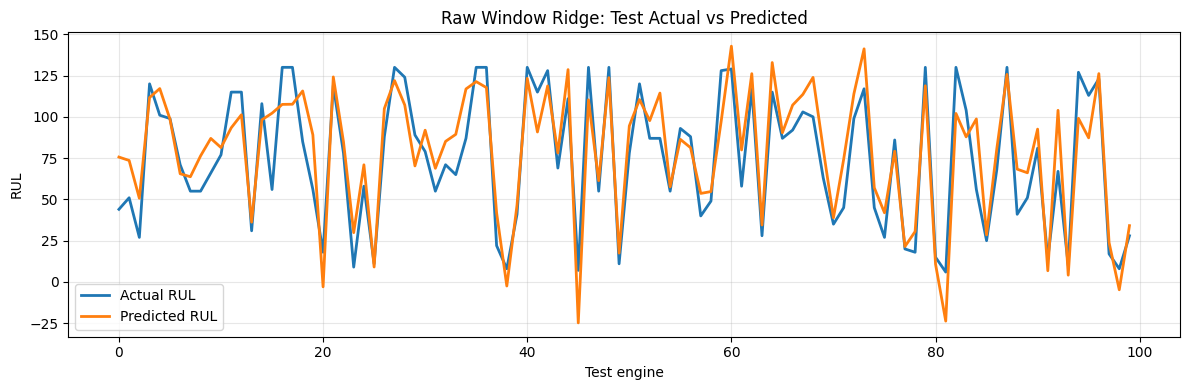}
    \caption{Predicted versus true RUL for Raw Ridge model on the FD001 (left) and FD003 (right) test sets.}
    \label{fig:predictions_raw_ridge}
\end{figure*}

\begin{figure*}[t]
    \centering
    \includegraphics[width=0.48\textwidth]{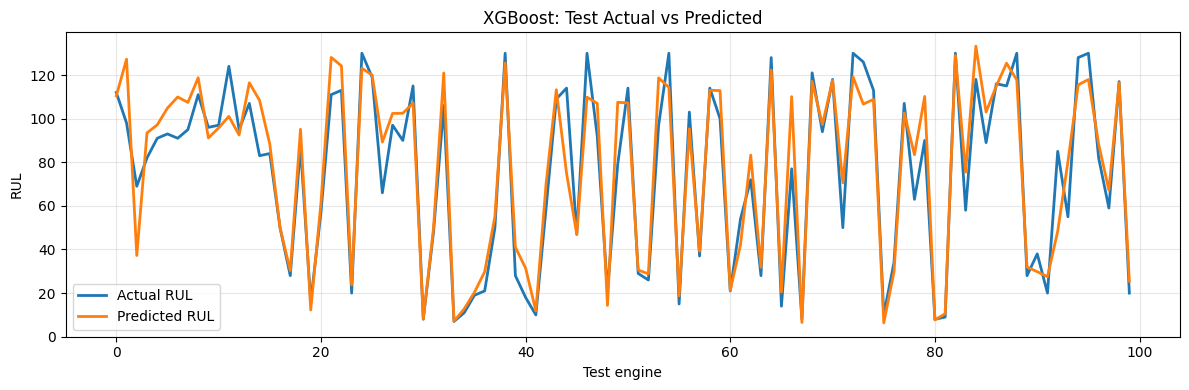}
    \hfill
    \includegraphics[width=0.48\textwidth]{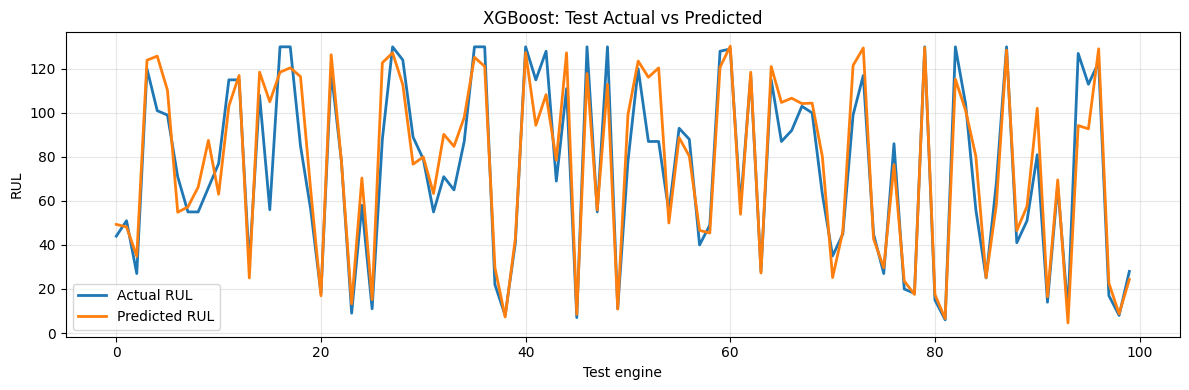}
    \caption{Predicted versus true RUL for XGBoost on the FD001 (left) and FD003 (right) test sets.}
    \label{fig:predictions_xgboost}
\end{figure*}

Figures~\ref{fig:predictions_raw_ridge} and~\ref{fig:predictions_xgboost} show predicted versus true RUL for the classical models. Raw Ridge exhibits higher variance and weaker alignment, while XGBoost achieves the closest tracking overall, particularly in mid-to-late degradation stages, with some deviations at extreme values.

\subsection{Scatter Analysis}

Although scatter plots are omitted for brevity, general trends are consistent across models and datasets. Prediction variance is higher at early life stages (high RUL values) and decreases as the engine approaches failure. This behavior reflects the increasing strength of degradation signals near end-of-life. Also, occasional late-prediction outliers, where RUL is overestimated near failure, can disproportionately inflate the NASA score, as it applies an exponential penalty to such optimistic errors, even when most predictions remain accurate.

\subsection{Temporal Learning Analysis}

To verify that the LSTM is genuinely learning temporal degradation patterns rather than relying solely on the most recent sensor readings, we conduct two analyses: hidden state visualization and sequence length sensitivity.

\begin{figure*}[t]
    \centering
    \includegraphics[width=0.48\textwidth]{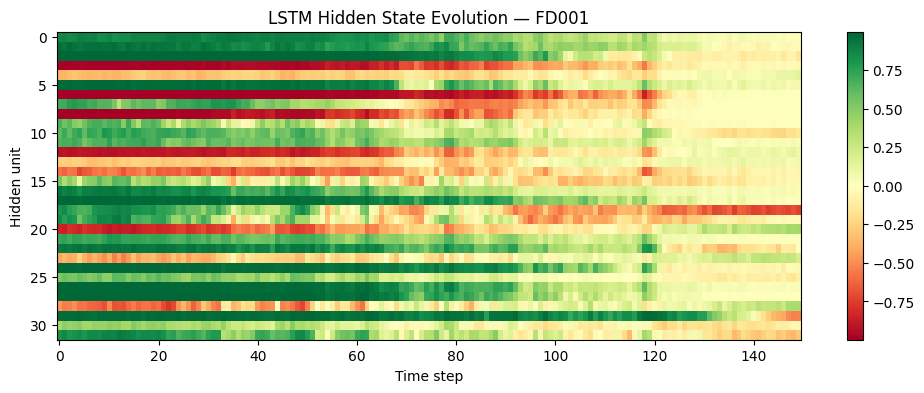}
    \hfill
    \includegraphics[width=0.48\textwidth]{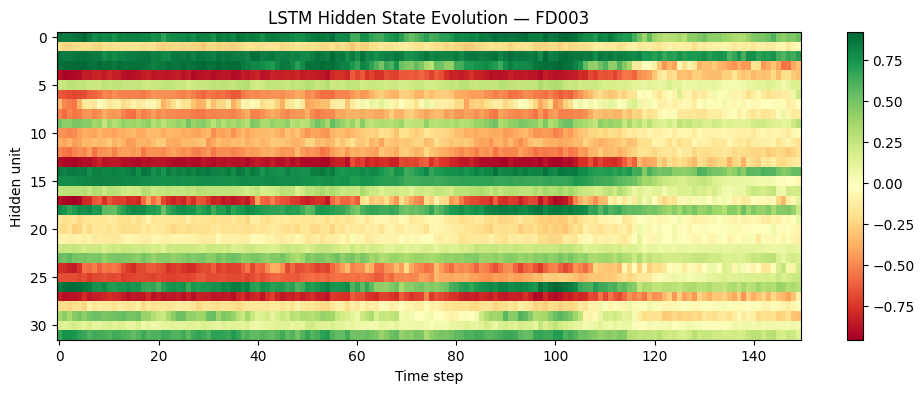}
    \caption{LSTM hidden state evolution for a single engine trajectory on FD001 (left) and FD003 (right). FD001 units show clear directional transitions toward failure while FD003 units exhibit more varied patterns reflecting its two fault conditions.}
    \label{fig:hidden_states}
\end{figure*}

Figure~\ref{fig:hidden_states} shows the hidden state activations of all 32 units across 150 consecutive windows for a single engine. On FD001, several units show clear directional transitions from healthy to degraded states as the engine approaches failure. On FD003, units display more complex and varied activation patterns, consistent with its two fault conditions.

Table~\ref{tab:seq_length} shows RMSE as input sequences are progressively shortened at test time. Both datasets show increasing RMSE as historical steps are removed, with FD003 degrading far more steeply --- removing just 5 steps increases RMSE by 48\% (14.20 to 21.06).

\begin{table}[h]
\caption{Effect of Sequence Length on LSTM Test RMSE}
\label{tab:seq_length}
\centering
\begin{tabular}{ccc}
\toprule
\textbf{Steps Removed} & \textbf{FD001 RMSE} & \textbf{FD003 RMSE} \\
\midrule
0  & 14.931 & 14.204 \\
5  & 15.978 & 21.060 \\
10 & 18.822 & 26.779 \\
15 & 20.059 & 27.372 \\
\bottomrule
\end{tabular}
\end{table}

\section{Discussion}

\subsection{Overall Comparison}

All models achieve meaningful RUL prediction on both datasets. The LSTM achieves the best RMSE and R\textsuperscript{2} on both FD001 and FD003, demonstrating the advantage of modeling temporal dependencies. The CNN achieves the lowest NASA score on FD001 (365 vs. 411 for LSTM), indicating more conservative predictions on the simpler single-fault dataset. Among models using engineered features, Ridge Regression provides a reasonable baseline, while XGBoost achieves the strongest performance, particularly on FD003.

\subsection{Model Analysis}

\textbf{Ridge Regression} benefits from slope and delta features that explicitly encode sensor degradation direction. However, its linear nature limits its performance on FD003, where multiple fault conditions produce heterogeneous sensor trajectories, reflected in its higher RMSE (17.98) compared to the CNN (15.68) and LSTM (14.20). Within the feature engineering approach, XGBoost (RMSE 13.36, NASA 396.7 on FD003) achieves the strongest performance, outperforming the CNN and approaching LSTM results. This highlights that nonlinear models on engineered features can effectively capture complex degradation patterns. The LSTM's advantage is that it achieves strong performance without requiring any hand-crafted features.

\textbf{1D CNN} efficiently captures local temporal patterns through convolution and achieves the lowest NASA score on FD001 (365), outperforming both Ridge Regression (469) and LSTM (411). This indicates that the CNN's local feature extraction produces more conservative and safety-aligned predictions on the single-fault FD001 dataset, where degradation patterns are consistent across engines. On FD003 however, the CNN's NASA score (648) is considerably worse than the LSTM (409) and even Ridge Regression (622). The scatter plot analysis reveals that a small number of late predictions on engines with atypical degradation trajectories, driven by the second fault mode, disproportionately inflate the asymmetric NASA score. This confirms that local convolutional feature extraction, while sufficient for uniform single-fault degradation, is insufficient for capturing the varied long-range dependencies introduced by multiple fault modes.

\textbf{LSTM} leverages its gating mechanism to retain relevant degradation history across the full sequence. It achieves the best RMSE on both datasets and the best NASA score on FD003, producing safer and more consistent predictions where degradation patterns are more varied. The sequence length sensitivity analysis further confirms that the LSTM's advantage on FD003 is specifically tied to its ability to integrate long-range temporal context; removing just 5 historical steps degrades FD003 RMSE by 48\%, far more than the 7\% degradation seen on FD001.

\subsection{Comparison with Prior Work}

Our LSTM outperforms Zheng et al.~\cite{paper} on both FD001 (RMSE 14.93 vs. 16.14) and FD003 (RMSE 14.20 vs. 16.18) using a simpler single-layer architecture. Two factors drive this improvement: the addition of a learning rate scheduler absent in the original paper, and an architectural search that identified a single LSTM layer as sufficient for single-operating-condition datasets.

\subsection{Asymmetric Cost of Errors}

The NASA score reveals important nuances beyond RMSE. On FD001, the CNN achieves the lowest score (365), indicating more conservative predictions than the LSTM (411) and Ridge Regression (469). This is a practically significant finding. In safety-critical applications, the CNN may be preferable on simple single-fault datasets where its local pattern extraction aligns well with consistent degradation signatures. On FD003, the LSTM score (409) is substantially better than the CNN (648) and Ridge Regression (619), demonstrating that temporal memory becomes critical when degradation patterns are heterogeneous. In such cases, the LSTM's ability to track gradual state transitions across the full sequence prevents the late predictions that disproportionately inflate the asymmetric NASA score.

\subsection{Limitations and Future Work}

This study covers FD001 and FD003, both with a single operating condition. The FD002 and FD004 subsets introduce six operating conditions, requiring regime normalization and likely benefiting deeper or hybrid architectures. Extending this comparison to FD002 and FD004 is the natural next step. Hybrid CNN-LSTM architectures~\cite{paper} and Transformer-based models also represent promising future directions. For the CNN specifically, per-regime normalization on FD003 may reduce the regime-driven variance that currently inflates late predictions and the NASA score.

\section{Conclusion}

We compared linear baselines (Ridge Regression, Polynomial Ridge, XGBoost), 1D CNN, and LSTM for turbofan engine RUL estimation on NASA C-MAPSS FD001 and FD003 under a fair and identical experimental pipeline. The LSTM achieves the best results among raw-sequence models (RMSE 14.93 on FD001, 14.20 on FD003), outperforming the deeper LSTM of Zheng et al.~\cite{paper} with a simpler single-layer design and a learning rate scheduler. The 1D CNN achieves RMSE of 16.97 and 15.68 on FD001 and FD003 respectively, and notably produces the most conservative and safety-aligned predictions on FD001 with the lowest NASA score (365) among all models on that dataset. This finding highlights a key tradeoff: CNN local feature extraction is well-suited to uniform single-fault degradation but is outperformed by LSTM when multiple fault modes introduce long-range temporal dependencies. Notably, extended experiments with engineered features show that XGBoost (RMSE 13.36, NASA 396.7 on FD003) can match or exceed deep learning performance, indicating that feature engineering and deep sequence modeling are complementary rather than competing approaches. Hidden state analysis and sequence length experiments confirm that the LSTM genuinely leverages temporal context, with FD003 showing particularly strong sensitivity to sequence history. Future work should extend this comparison to FD002 and FD004, and investigate per-regime normalization as a preprocessing improvement for multi-condition subsets.

\section*{Acknowledgements}

We would like to thank Dr. Jerome J. Braun for his guidance and for providing us with the knowledge necessary to carry this project forward.

\balance


\begin{thebibliography}{9}

\bibitem{paper}
S. Zheng, K. Ristovski, A. Farahat, and C. Gupta, ``Long Short-Term Memory Network for Remaining Useful Life Estimation,'' in \textit{Proc. IEEE Int. Conf. Prognostics and Health Management (ICPHM)}, 2017, pp. 88--95.

\bibitem{cmapss}
A. Saxena and K. Goebel, ``Turbofan Engine Degradation Simulation Data Set,'' NASA Ames Research Center, Moffett Field, CA, USA, Tech. Rep., 2008.

\bibitem{lstm}
S. Hochreiter and J. Schmidhuber, ``Long Short-Term Memory,'' \textit{Neural Computation}, vol. 9, no. 8, pp. 1735--1780, 1997.

\bibitem{bengio1994}
Y. Bengio, P. Simard, and P. Frasconi, ``Learning Long-Term Dependencies with Gradient Descent is Difficult,'' \textit{IEEE Trans. Neural Networks}, vol. 5, no. 2, pp. 157--166, 1994.

\bibitem{babu2016}
G. S. Babu, P. Zhao, and X.-L. Li, ``Deep Convolutional Neural Network Based Regression Approach for Estimation of Remaining Useful Life,'' in \textit{Proc. Int. Conf. Database Systems for Advanced Applications}, Springer, 2016, pp. 214--228.

\end{thebibliography}
\end{document}